\pdfoutput=1

\documentclass[11pt]{article}

\usepackage{acl}

\usepackage{times}
\usepackage{latexsym}
\usepackage{graphicx}
\usepackage{arabtex}
\usepackage{utf8}
\setcode{utf8}
\usepackage{amsmath}
\usepackage{algpseudocode}
\usepackage{algorithm}
\usepackage{enumitem}
\usepackage{multirow}
\usepackage{xcolor,colortbl,soul}
\usepackage{array,multirow,booktabs}

\usepackage[T1]{fontenc}

\usepackage[utf8]{inputenc}

\usepackage{microtype}

\definecolor{LightCyan}{rgb}{0.88,1,1}
\definecolor{LightPink}{rgb}{0.98,0.8,0.91}

\title{Syntax and Semantics Meet in the ``Middle'': \\ Probing the Syntax-Semantics Interface of LMs Through Agentivity}

\author{Lindia Tjuatja, Emmy Liu, Lori Levin, Graham Neubig \\
    Language Technologies Institute\\
    Carnegie Mellon University\\
  \texttt{\{ltjuatja, mengyan3, lsl, gneubig\}@cs.cmu.edu} \\\
}

\begin{document}
\maketitle

\begin{abstract}
Recent advances in large language models have prompted researchers to examine their abilities across a variety of linguistic tasks, but little has been done to investigate how models handle the interactions in meaning across words and larger syntactic forms---i.e. phenomena at the intersection of syntax and semantics. We present the semantic notion of {\it agentivity} as a case study for probing such interactions. We created a novel evaluation dataset by utilitizing the unique linguistic properties of a subset of optionally transitive English verbs. This dataset was used to prompt varying sizes of three model classes to see if they are sensitive to agentivity at the lexical level, and if they can appropriately employ these word-level priors given a specific syntactic context. Overall, GPT-3 \texttt{text-davinci-003} performs extremely well across all experiments, outperforming all other models tested by far. In fact, the results are even better correlated with human judgements than both syntactic and semantic corpus statistics. This suggests that LMs may potentially serve as more useful tools for linguistic annotation, theory testing, and discovery than select corpora for certain tasks.\footnote{Code is available at \url{https://github.com/lindiatjuatja/lm_sem}} 
\end{list} 
\end{abstract}

\section{Introduction} \label{section:intro}
Consider the English sentences in (\ref{ex:write}) below:
\begin{enumerate}
\item 
    \label{ex:write}
        \begin{enumerate}[label=\alph*., ref=\alph*]
        \itemsep0em 
            \item This author writes easily.
                \label{ex:write-a}
            \item This passage writes easily.
                \label{ex:write-b}
        \end{enumerate}
\end{enumerate}
These sentences display an interesting property of certain optionally transitive verbs in English. Although they share an identical surface syntactic structure---a noun phrase in subject position followed by the intransitive form of the verb and an adverb phrase modifying the verb---they entail very different things about the roles of their subjects. 

The subject of (\ref{ex:write}\ref{ex:write-a}) is someone that does the action of writing; in other words, {\it this author} is an \textbf{agent} in the writing event. On the other hand, the subject of (\ref{ex:write}\ref{ex:write-b}), {\it this passage}, doesn't do any writing---it is what is created in the event of writing. In contrast to {\it this author}, {\it this passage} is a \textbf{patient}. The agent and patient roles are not discrete categories, but rather prototypes on opposite ends of a continuum. These ``protoroles'' have a number of contributing properties such as causing an event for agents and undergoing change of state for patients \cite{dowty1991thematic}.

The contrast between the minimal pair in (\ref{ex:write}) suggests that there are lexical semantic properties of the subjects that give rise to these two distinct readings: one that describes how the subject generally {\it does} an action as in (\ref{ex:write}\ref{ex:write-a}), and another that describes how an event generally unfolds when the subject {\it undergoes} an action as in (\ref{ex:write}\ref{ex:write-b}). 
Intuitively, a speaker may know from the meaning of {\it author} that authors are animate, have some degree of volition, and typically write things, whereas passages (of text) are inanimate, have no volition, and are typically written. The knowledge of these aspects of meaning must somehow interact with the syntactic form of the sentences in (\ref{ex:write}) to disambiguate between the two possible readings, and an agent or patient role for the subject follows from the meaning of the statement as a whole.

Now consider the (somewhat unusual) sentences in (\ref{ex:write-trans}) which use the transitive form of {\it write}:
\begin{enumerate}[resume]
\item 
    \label{ex:write-trans}
        \begin{enumerate}[label=\alph*., ref=\alph*]
        \itemsep0em 
            \item Something writes this author easily.
                \label{ex:trans-a}
            \item This passage writes something easily.
                \label{ex:trans-b}
        \end{enumerate}
\end{enumerate}
At first glance, the above sentences (with the same sense of {\it write} as in \ref{ex:write}) are infelicitous unless we imagine some obscure context where {\it this author} is something like a character in a text and {\it this passage} is somehow anthropomorphized and capable of writing; these contexts go against our natural intuitions of the semantics of ``passage'' and ``author''.\footnote{There is another reading of (\ref{ex:write-trans}\ref{ex:trans-a}) that uses a different sense of {\it write}, where {\it this author} is a recipient ({\it Something writes (to) this author easily}). Regardless, given that the agent and patient roles as defined by \citet{dowty1991thematic} are prototypes on a scale, {\it this author} in the recipient reading is closer to the patient role.} Unlike the syntactic form of the sentences in (\ref{ex:write}), the explicit inclusion of both arguments (subject and direct object) now forces whatever is in subject position to be the agent and whatever is in object position to be more like a patient, regardless of the typical semantic properties of the arguments.

Taken together, the examples in (\ref{ex:write}) and (\ref{ex:write-trans}) illustrate a compelling interaction at the {\it syntax-semantics interface}. More specifically, we see a two-way interaction: first, near-identical surface forms acquire completely different entailments about their subjects {\it solely} depending on the choice of subject, while conversely certain syntactic forms can influence the semantic role of an argument {\it regardless} of the usual behavior of said argument. We aim to investigate the linguistic capabilities of language models with regards to this interaction.

Prior work in studying LMs as psycholinguistic subjects has largely focused on syntax and grammatical well-formedness (\citealt{futrell-etal-2019-neural, Linzen_2021}, inter alia). However, as illustrated in the above examples, there are instances of near-identical syntactic structures that can give rise to different meanings depending on the individual lexical items as well as surrounding context. Thus evaluating LMs on syntax, while a necessary starting point, does not give us a sufficient measure of LM linguistic capabilities. While other work such as \citet{ettinger-2020-bert}, \citet{kim-linzen-2020-cogs}, and \citet{misra2022comps} (among others) evaluate LMs on a variety of tests involving semantics and pragmatics, they do not investigate the interaction between the meanings associated with syntactic forms and those of individual lexical items.

Thus, we not only need to evaluate syntax and utilization of semantic knowledge, but we also need to understand how interactions of meaning at different linguistic levels---i.e. morphological, lexical, phrasal---may alter model behavior. Exploring phenomena within the syntax-semantics interface is a compelling approach as it gives us access to specific aspects of semantics while allowing precise control over syntactic form between levels.

In this work, we probe the syntax-semantics interface of several language models, focusing on the semantic notion of agentivity. We do this by prompting models to label nouns in isolation or in context as either agents or patients from a curated test set of noun-verb-adverb combinations that display the alternation shown in example (\ref{ex:write}). We then compare the performance of LMs to both human judgements and corpus statistics. 

Probing for LMs for their knowledge of agentivity in syntactic constructions as in (\ref{ex:write}) and (\ref{ex:write-trans}) is a particularly insightful case study as it allows us to explore three interconnected questions in a highly controlled syntactic setting:
\begin{enumerate}[label=\Roman*.]
    \item Do models display sensitivity to aspects of word-level semantics independent of syntactic context, and is such sensitivity aligned with human judgements? (\S\ref{section:exp1}) 
    \item Can models employ lexical semantics to determine the appropriate semantics of a sentence where the syntax is ambiguous between readings (as in \ref{ex:write})? (\S\ref{section:exp2})
    \item Can models determine the semantics of a sentence from syntax, disregarding lexical semantics when necessary (as in \ref{ex:write-trans})? (\S\ref{section:exp3})
\end{enumerate}

Additionally, the relatively infrequent pairings of semantic function and syntactic form of sentences such as (1\ref{ex:write-b}) are also interesting from a learnability and acquisition perspective for both LMs and humans. How both come to process and acquire exceptions to a general ``rule'' has been a topic of debate since early connectionist models \cite{rumelhart1986learning}. Hence, knowledge of LM capabilities in acquiring and processing these linguistic anomalies may serve as valuable insight to linguists, cognitive scientists, and NLP practitioners alike.

\section{Methodology}
We constructed three experiments, each targeting one of the above questions through the lens of agentivity. We will first give a broad overview of each, and then go into detail about the general approach. 

\begin{figure*}[t]
\centering
\includegraphics[width=\textwidth]{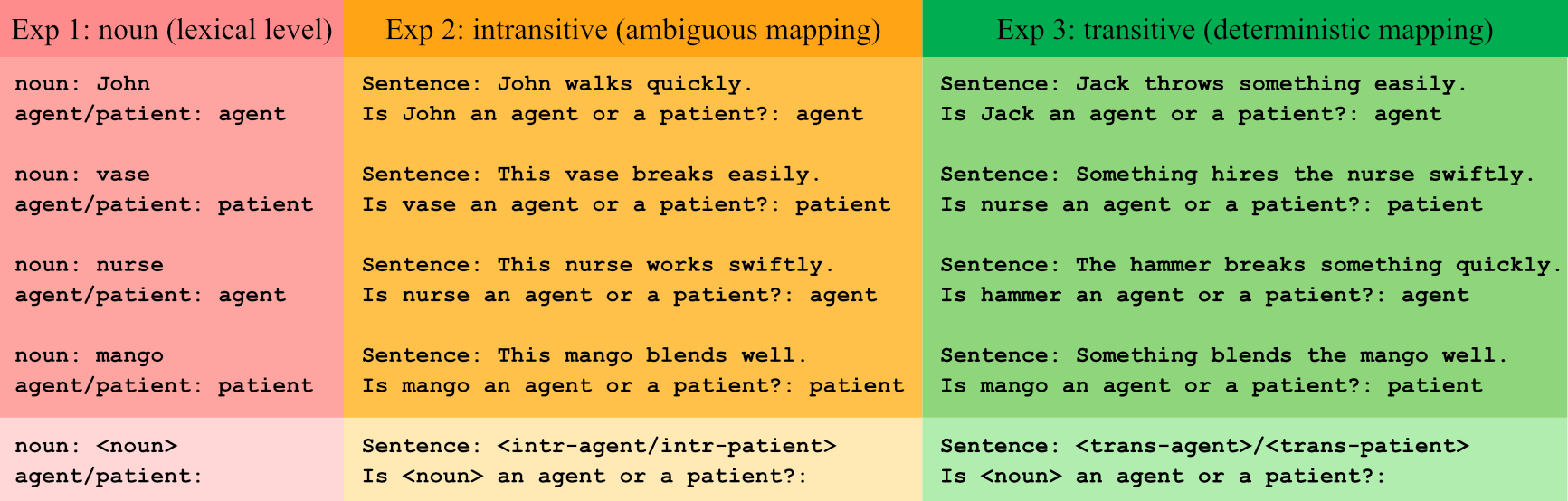}
\caption{\label{fig:prompt} Prompt setup for each experiment. Note that the examples given for Exp 1 are not meant to be hard labels, rather they are ``tendencies'' for these nouns. In Exp 2, the noun itself determines whether the sentence is considered \textbf{intr-agent} or \textbf{intr-patient}; in Exp 3, we force the noun to take the agent or patient role by placing it in subject (\textbf{trans-agent}) or object (\textbf{trans-patient}) position.
}
\end{figure*}

\textbf{Experiment 1} (\S\ref{section:exp1}) tests whether language models are sensitive to the word-level semantics of nouns with regards to agentivity, such as whether nouns like {\it author} and {\it passage} are more likely to be agents or patients without any surrounding context. This is analogous to the idea that speakers have intuition for how entities prototypically act in events, e.g. that {\it authors} write and {\it passages} are written, and that this extends to how we categorize their roles in events \cite{rissman2019thematic}. 

\textbf{Experiment 2} (\S\ref{section:exp2}) tests whether language models can disambiguate between the possible readings of sentences of the form in (\ref{ex:write})---i.e. if they can identify whether the syntactic subject is an agent or a patient when the verb can allow for either. Sentences with the intransitive form of the verb that describe how the subject (an agent) does an action demonstrate {\it object drop} (as the direct object of the normally transitive verb is ``dropped''), while sentences that describe how an event unfolds when the subject (a patient) undergoes an action are called {\it middles}, short for the linguistic term {\it dispositional middle} (\citealt{Oosten77, Jaeggli86, Condoravdi89, Fagan92}, inter alia).\footnote{Note that in English, dispositional middles also allow for what are considered non-patient promoted objects (such as paths, e.g. {\it The desert crosses easily}) (\citealt{Tenny94, Tenny92-cr}), but for convenience we will treat them as being in the same category as patients.} In our experimental setup, we will refer to these as \textbf{intr-agent} and \textbf{intr-patient}, respectively. If a model can do this task successfully by employing semantic information about the noun, we would expect not only to see that nouns in subject position are classified correctly as agents or patients, but also that these predictions for the most part correlate to the predictions in the first experiment. 

Finally, \textbf{Experiment 3} (\S\ref{section:exp3}) tests whether language models can disregard word-specific priors to identify whether the noun of interest in a sentence with a transitive verb (such as those in \ref{ex:write-trans}) is an agent or patient. Since the semantic role of the noun maps directly to its syntactic position in these sentences, all subjects should be agents and all objects should be patients. For our test set, we create sentences where the position of the noun is the subject (\textbf{trans-agent}) and sentences where it is the object (\textbf{trans-patient}) for every noun. 

\subsection{General approach and data curation}
In all of these experiments, we rely on the prompting paradigm to elicit LM probabilities of an ``agent'' or ``patient'' label for a given noun in isolation or within a sentence. Our prompting method consists of four examples with gold labels, followed by the unlabeled test example in the same format, as shown in Figure \ref{fig:prompt}. As this task has not been explored in prior literature, we had to construct our own examples to test on. 

The highly controlled syntactic setting that allows us to explore the alternation in agentivity as displayed in (\ref{ex:write}) and (\ref{ex:write-trans}) is a double-edged sword---while this setting provides us with a minimal pair, it also restricts the types of verbs that work in this experimental setup. The second (\textbf{intr-agent} vs. \textbf{intr-patient}) and third (\textbf{trans-agent} and \textbf{trans-patient}) experiments require verbs that are optionally transitive and have no preference for whether an agent or a patient is the subject of the intransitive form, as in (\ref{ex:write}). These requirements together highly constrain the class of verbs that work in this experimental setup, and as far we can tell there exists no definitive list in the linguistics literature of English verbs that display both properties. 

As a starting point to curate a list of verbs, we consulted literature on verbs that display object drop (\citealt{gillon2012implicit, fillmore1986pragmatically}, as well as \citealt{levin1993english} for an overview of English verb classes). We compiled a list of 23 verbs (see Appendix \ref{appendix:noun-verb-adv}), though this list is certainly non-exhaustive. For each verb, we list nouns and adverbs that can work in combination with each other in all of the \sethlcolor{lightgray}\hl{templates} in Table \ref{tab:templates}. Criteria for adding nouns and adverbs are listed in the Appendix \ref{appendix:data-curation}. 

In total, we have 233 unique nouns and a total of 820 noun-verb-adverb combinations. Out of these combinations, 343 form \textbf{intr-agent} sentences and 477 form \textbf{intr-patient} sentences. Since we can put any noun into syntactic subject or object position for the transitive sentences, we have 820 sentences each for \textbf{trans-agent} and \textbf{trans-patient}.

\begin{table}[h]
\small
\centering
\begin{tabular}{cl}
\toprule
\textbf{Sentence} & \multicolumn{1}{c}{\textbf{Template}} \\
\midrule
& \cellcolor{lightgray} This \textbf{<noun> <verb>} \textbf{<adverb>}. \\
\textbf{intr-agent} & \cellcolor{LightPink} This author writes easily. \\
\textbf{intr-patient} & \cellcolor{LightCyan}This paper writes easily. \\
\midrule \midrule
& \cellcolor{lightgray} This \textbf{<noun> <verb>} something \textbf{<adv>}. \\
 \multirow{2}{*}{\textbf{trans-agent}} & \cellcolor{LightPink} This author writes something easily. \\
& \cellcolor{LightPink} This paper writes something easily. \\
\arrayrulecolor{white}
\midrule
\arrayrulecolor{black}
& \cellcolor{lightgray} Something \textbf{<verb>} this \textbf{<noun> <adv>}. \\
\multirow{2}{*}{\textbf{trans-patient}} & \cellcolor{LightCyan} Something writes this author easily. \\
& \cellcolor{LightCyan} Something writes this paper easily. \\
\bottomrule
\end{tabular}
\caption {Templates for experiments 2 and 3. Sentences highlighted in \sethlcolor{LightPink}\hl{pink} contain a \textbf{<noun>} with an ``agent'' label, while those in \sethlcolor{LightCyan}\hl{blue} with ``patient''.}
\label{tab:templates}
\end{table}

\subsection{Approximating ``ground truth'' agentivity labels for nouns out of context}

Getting a gold ``agent'' or ``patient'' label is straightforward in the experiments with nouns in context: for sentences with the intransitive this was done ad hoc during data curation, and for sentences with the transitive this is a one-to-one mapping to syntax. However, using a hard label for nouns in isolation is problematic as a semantic role label is meaningless without context of the event; in principle, given an appropriate context, anything can act upon something else or have something done to it (literally or figuratively). 

To get around this, we have two methods for finding an approximate label for the ``typical'' agentivity of a noun. The first was to collect human judgements. 19 annotators (native/fluent bilingual English proficiency) were given nouns without any context and were tasked to judge how likely each noun is to be an agent in any arbitrary event where both an agent and patient are involved. Their judgements were collected via ratings on a scale from 1 (very unlikely to be an agent) to 5 (very likely to be an agent). For nouns that have multiple common word senses (e.g. ``model'' can refer to both a fashion model or machine learning model, among other things) we include a disambiguating description. This description does not contain any verbs or other explicit indications of what events the noun may occur in (e.g. for ``model'', we give human annotators ``model (person)'').\footnote{Additional details on collecting human ratings can be found in Appendix \ref{appendix:human}.} We then average the ratings across all annotators and normalize so that the values fall between 0 and 1. To calculate inter-annotator agreement, we randomly divide the annotators into two groups (of 9 and 10), average their ratings for each noun, and calculate the correlation between the two; doing this seven times yields an average inter-group correlation of 0.968.

The second method uses statistics from linguistically annotated corpora as a proxy for the ``typical'' agentivity of a noun. We do this by calculating the frequency of ``agenthood'' for a noun (\textbf{agent ratio}), i.e. dividing the number of times the noun appears as an agent by the number of times it is either an agent or patient. The ideal annotated corpus for this would be one with semantic role labels such as Propbank \cite{kingsbury2002treebank}, where the ``ARG0'' label corresponds to agent and ``ARG1'' to patient. However, many of the nouns in our data appeared only a few times in Propbank or not at all---out of all 233 nouns, only 166 of them occurred within an ARG0 or ARG1 span.\footnote{We used Propbank annotations for BOLT, EWT, and Ontonotes 5.0 from \url{https://github.com/propbank/propbank-release}.} 

Thus, we also tried utilizing syntax as a proxy using Google Syntactic Ngrams biarcs \cite{goldberg-2013-dataset}, as it is significantly larger. The biarcs portion of the corpus covers dependency relations between three connected content words, which includes transitive predicates. To calculate a similar ratio, we divide the number of times a noun occurs as a subject by the total number of subject and direct object occurrences (we call this the \textbf{subject ratio}). A value closer to 1 should correlate with a tendency to occur more often as an agent, as agents are generally coded as subjects of English transitive verbs and patients as direct objects. All but one of our nouns contained at least one instance of occurring with a ``nsubj'' or ``dobj'' label.

\begin{figure}
\centering
\includegraphics[width=\linewidth]{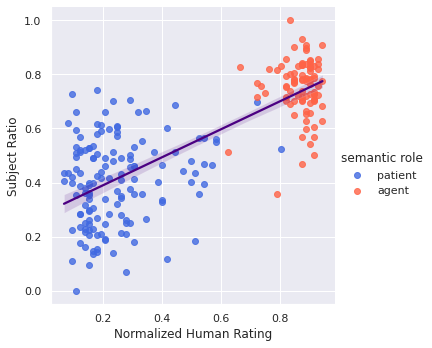}
\caption{\label{fig:human-ngrams} Correlation between subject ratio (from Google Syntactic Ngrams) and human ratings for each noun ($r=0.762$). The semantic role label is the role the noun takes as the subject of the intransitive verb within our test set.
}
\end{figure}

\section{Experimental Results} 
We evaluate BLOOM \cite{scao2022bloom}, GPT-2 \cite{radford2019language}, and GPT-3 \cite{brown2020language} models of varying sizes for all experiments. Since previous work has shown that models are highly sensitive to the ordering of examples \cite{lu2021prompting}, we run each experiment twice: once with the order shown in Figure \ref{fig:prompt} where an agent is first (APAP ordering) and again with the first example moved to the bottom (PAPA ordering). We compare models based on their average performance across both orderings. Note, however, that some models are more sensitive to orderings than others; some models (like \texttt{text-davinci-003}) are largely invariant to example ordering. In Appendix \ref{appendix:order}, we report results from both experiments. 

\subsection{Exp 1: Agentivity at the lexical level} \label{section:exp1}
In order to see if models are sensitive to the notion of how ``typically'' agentive a noun is, we compare the difference in log-likelihood between predicting ``agent'' or ``patient'' for that noun ($\delta$-LL) with the normalized human ratings as well as corpus statistics from Google Syntactic Ngrams and Propbank. 

Before we compare models with Ngrams and Propbank, we first ask how well-correlated both are with human ratings. We find that the subject ratio calculated from occurrence counts in Google Syntactic Ngrams is positively correlated with the average human rating with Pearson's $r$ of 0.762, though the human rating has a stronger divide between agents and patients. This can be seen in Figure \ref{fig:human-ngrams}. When comparing with humans, using Syntactic Ngrams for this task actually turns out to be better than using Propbank: for the 166 nouns that occur with ARG0/1 labels, there is a correlation of 0.555 with human ratings (see Appendix \ref{appendix:propbank} for details).

\begin{figure}[h]
\centering
\includegraphics[width=\linewidth]{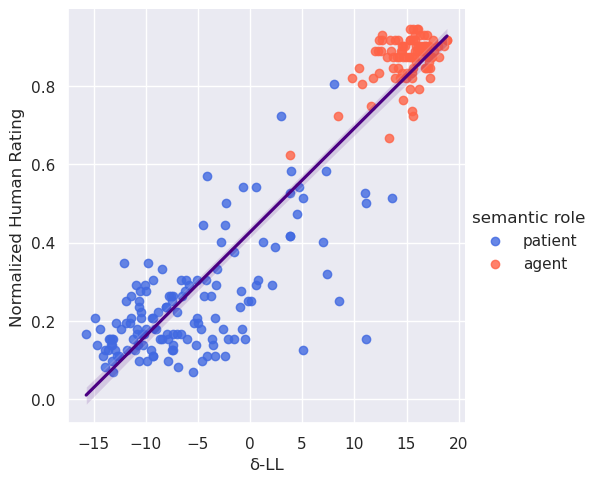}
\caption{\label{fig:noun_corr-davinci} Correlation between $\delta$-LL in Experiment 1 for GPT-3 \texttt{davinci-003} and the normalized human rating in the APAP experiment. Note that a negative $\delta$-LL means the ``patient'' label is more likely.}
\end{figure}

Overall, as seen in Table \ref{tab:noun-corr}, we find that most models have a weak correlation with human ratings, with the exception of GPT-3 \texttt{text-davinci-003} (henceforth \texttt{davinci-003}), shown in Figure \ref{fig:noun_corr-davinci}. We also see that \texttt{davinci-003} is not only both better correlated with human judgements than with corpus statistics, but surprisingly there is also a stronger correlation between its  $\delta$-LL and human ratings than between these proxies (syntactic and semantic) and human ratings. In fact, \texttt{davinci-003} is extremely close to the average inter-annotator group correlation, and furthermore this correlation is largely invariant to the ordering of prompts. 

The observation that \texttt{davinci-003} is better correlated with human judgement than both syntactic (Ngrams) and semantic (Propbank) corpus statistics is intriguing as both types of corpora have been used in modeling prediction of {\it thematic fit}, or how well a noun fulfills a certain thematic role with a verb \cite{sayeed-etal-2016-thematic}. Thus, we may naturally expect this to also work well with ``general tendencies'' or typicality judgements for nouns by themselves. However, it seems that such corpora may be too small or genre-biased to fully capture the nuances of human judgements, and such judgements may be better captured by LMs that have seen vast quantities of data across a wide variety of domains, even without explicit human annotation.

\begin{table}[h]
\small
\centering
\resizebox{\columnwidth}{!}{%
\begin{tabular}{lccc}
\toprule
\textbf{Model} & \textbf{Human} & \textbf{Ngrams} & \textbf{PB} \\
\midrule
BLOOM \texttt{560m} & 0.549 & 0.519 & 0.377 \\
BLOOM \texttt{1b1} & 0.374 & 0.358 & 0.291 \\
BLOOM \texttt{1b7} & 0.340 & 0.288 & 0.278 \\
BLOOM \texttt{3b} & 0.305 & 0.348 & 0.231 \\
BLOOM \texttt{7b1} & 0.016 & -0.129 & 0.011 \\
\midrule
GPT-2 \texttt{small} & 0.650 & 0.569 & 0.463 \\
GPT-2 \texttt{medium} & 0.394 & 0.451 & 0.333 \\
GPT-2 \texttt{large} & 0.499 & 0.544 & 0.412 \\
GPT-2 \texttt{xl} & 0.358 & 0.349 & 0.227 \\
\midrule
GPT-3 \texttt{ada-001} & 0.594 & 0.575 & 0.490 \\
GPT-3 \texttt{babbage-001} & 0.311 & 0.337 & 0.158 \\
GPT-3 \texttt{curie-001} & 0.107 & 0.181 & 0.128 \\
GPT-3 \texttt{davinci-001} & 0.467 & 0.461 & 0.330 \\
GPT-3 \texttt{davinci-003} & \textbf{0.939} & \textbf{0.730} & \textbf{0.574} \\
\midrule \midrule
Inter-annotator & 0.968 & -- & -- \\
Google Syntactic Ngrams & 0.762 & -- & -- \\
Propbank & 0.555 & -- & -- \\
\bottomrule
\end{tabular}%
}
\caption{Correlation between the difference in log-likelihood of predicting ``agent'' or ``patient'' with human ratings, subject ratio calculated from Google Syntactic Ngrams (232/233 nouns), and agent ratio calculated from Propbank (166/233 nouns), averaged across APAP and PAPA experiments.} 
\label{tab:noun-corr}
\end{table}

\subsection{Exp 2: Disambiguating agentivity with the intransitive} \label{section:exp2}
In this experiment, we evaluate models along two metrics: how accurate the model is in predicting the correct label in context and how strongly correlated the $\delta$-LL in this experiment is with the $\delta$-LL from Experiment 1. 

\begin{figure}[h]
\centering
\includegraphics[width=\linewidth]{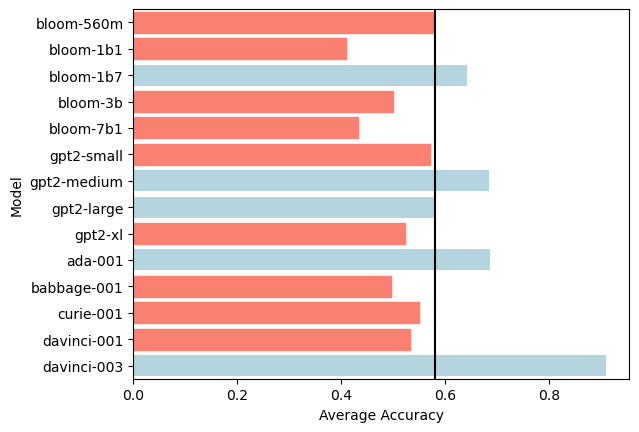}
\caption{\label{fig:intr_avg_acc} Average accuracy for predicting the label of nouns in \textbf{intr-agent}/\textbf{intr-patient} sentences. The black line indicates majority class performance; blue bars indicate above majority class performance. 
}
\end{figure}

Figure \ref{fig:intr_avg_acc} shows the accuracy of each model in predicting (giving a higher probability to) the correct semantic label. Over half of the models do not achieve chance performance (predicting the majority class $\approx 0.582$). Interestingly, we find that there is no monotonic increase in performance for this task with respect to model size \cite{kaplan2020scaling}---for example, performance drops drastically between \texttt{text-ada-001} and \texttt{text-babbage-001}. This is also the case in Experiment 1. 

We also evaluate how strongly correlated the $\delta$-LL between predicting ``agent'' or ``patient'' for the noun in subject position of the intransitive is with the $\delta$-LL of the noun in isolation. Since the role of the noun in the intransitive is heavily dependent on the meaning of the noun itself, if a model is using this information to disambiguate we would expect that the $\delta$-LL in this experiment is correlated with $\delta$-LL from Experiment 1. Furthermore, we would also want it to be strongly correlated with our approximate ``ground truth'' measures for agentivity, especially human ratings. 

These correlations are shown in Table \ref{tab:intr-corr}. As expected, \texttt{davinci-003} displays a strong relationship between the $\delta$-LL from intransitive sentences with the $\delta$-LL from Experiment 1, and furthermore also has a strong correlation with human ratings. Like in Experiment 1, \texttt{davinci-003}'s performance is invariant to changes in example orders.

\begin{table}[h]
\small
\centering
\resizebox{\columnwidth}{!}{%
\begin{tabular}{lcccc}
\toprule
\textbf{Model} & \textbf{Noun $\delta$-LL} & \textbf{Human} & \textbf{Ngrams} & \textbf{PB} \\
\midrule
BLOOM \texttt{560m} & 0.605 & 0.217 & 0.147 & 0.100 \\
BLOOM \texttt{1b1} & 0.702 & -0.0344 & 0.0200 & 0.0511\\
BLOOM \texttt{1b7} & 0.540 & 0.706 & 0.562 & 0.441 \\
BLOOM \texttt{3b} & 0.258 & 0.280 & 0.190 & 0.0871 \\
BLOOM \texttt{7b1} & 0.385 & 0.161 & 0.124 & 0.0689 \\
\midrule
GPT-2 \texttt{small} & 0.655 & 0.424 & 0.309 & 0.290 \\
GPT-2 \texttt{medium} & 0.611 & 0.523 & 0.516 & 0.505 \\
GPT-2 \texttt{large} & 0.551 & 0.609 & 0.489 & 0.447 \\
GPT-2 \texttt{xl} & 0.548 & 0.507 & 0.445 & 0.363 \\
\midrule
GPT-3 \texttt{ada-001} & 0.541 & 0.496 & 0.358 & 0.307 \\
GPT-3 \texttt{babbage-001} & 0.127 & -0.176 & -0.170 & -0.125 \\
GPT-3 \texttt{curie-001} & 0.130 & 0.156 & 0.189 & 0.0953 \\
GPT-3 \texttt{davinci-001} & 0.487 & 0.647 & 0.515 & 0.376 \\
GPT-3 \texttt{davinci-003} & \textbf{0.914} & \textbf{0.919} & \textbf{0.715} & \textbf{0.567}\\
\bottomrule
\end{tabular}%
}
\caption{Correlation between the $\delta$-LL from \textbf{intr-agent}/\textbf{intr-patient} sentences with the $\delta$-LL from the noun in isolation, human ratings, subject (Google Syntactic Ngrams), and agent ratios (Propbank).}
\label{tab:intr-corr}
\end{table}

\subsection{Exp 3: Agentivity with the transitive} 
\label{section:exp3}

\begin{figure}[h]
\centering
\includegraphics[width=\linewidth]{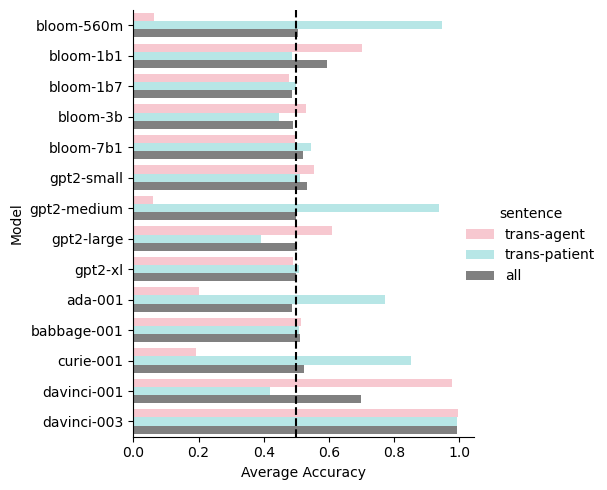}
\caption{\label{fig:trans_avg_acc} Average accuracy across \textbf{trans-agent}, \textbf{trans-patient}, and all transitive sentences. The dashed line indicates chance performance.}
\end{figure}

As previously discussed, the syntactic position of the noun in the transitive sentences (subject or object) directly map to their semantic roles (agent and patient, respectively). Figure \ref{fig:trans_avg_acc} shows accuracy split by \textbf{trans-agent} and \textbf{trans-patient}.

As in the previous experiments, GPT-3 \texttt{davinci-003} outperforms all other models (0.994 for \textbf{trans-agent} and 0.991 for \textbf{trans-patient}---it is actually the {\it only} model which performs significantly above chance for both Experiments 2 and 3, and is also consistent across both example orderings.

\section{A Closer Look at davinci-003}
Given that GPT-3 \texttt{davinci-003} does extremely well, a natural question to ask is whether \texttt{davinci-003} ``fails'' in similar ways to humans---i.e. we can see whether the nouns that are misclassified in the intransitive sentence setting (\S\ref{section:exp2}) are more ambiguous to humans as well. 

In both APAP and PAPA orderings, all or nearly all of what \texttt{davinci-003} gets incorrect are patient subjects; all 78 incorrectly classified subjects of sentences in the APAP ordering are patients, and 69 of the 70 incorrect subjects in the PAPA ordering are patients. From this, one way to answer the above question is to compare this subset of nouns with the subset of nouns with a ``patient'' label (in the intransitive construction) that humans tend to rate as more agentive. 

\subsection{Animacy and thematic fit}
\label{section:animacy-thematic}
Table \ref{tab:human-patient-nouns} lists the latter subset of nouns, i.e. the most ``agent-like'' nouns with a ``patient'' label in the intransitive construction. Recall that human annotators were asked to rate each noun in isolation from a scale from 1 (very unlikely to be an agent) to 5 (very likely to be an agent) which is then normalized to a scale from 0 to 1, whereas the gold labels for nouns are determined by role it takes in the constructed (in this case, intransitive) sentences. 

Animate nouns, such as ``model (person)'', ``animal'', and ``fish'' are unsurprisingly in this list, as many linguists have noted that the notion of agentivity is closely related to animacy (\citealt{silverstein1976shifters, comrie1989language}, inter alia). However, across both orderings, the only noun that was misclassified was ``model'' in the sentence {\it This model photographs beautifully/nicely}. Nevertheless, it could be argued that an agent interpretation in this context is plausible. 

It appears that there are two interactions that are occurring in the above example. First, we must consider the {\it selectional restrictions} and of the verb, i.e. what arguments are allowable in the event described by the verb (\citealt{chomsky1965aspects, katz1963structure}). While selectional restrictions are traditionally viewed as binary features, a weaker, gradient version of this is {\it selectional preferences}, or the degree to which an argument fulfills the restrictions of the event \cite{resnik1996selectional}. A closely related notion to this is {\it thematic fit}, which is how much a word fulfills these preferences.  

Secondly, the {\it Animacy Hierarchy}---of which humans are at the top---plays a role in such selectional restrictions and preferences, and thus in thematic fit \cite{trueswell-1994}. Since {\it photograph} requires a human-like entity as an agent, it could be argued that the interpretation of ``model'' being an agent in this sentence is not invalid (though likely a less salient interpretation by English speakers), as nothing in the ``photographing'' event rules out a subtype of a human ``model'' being the agent. This contrasts with the example with ``animal'' in our test set ({\it This animal photographs beautifully/nicely}), which would be far less acceptable with an animal agent interpretation, and falls below ``human model'' in the Animacy Hierarchy. 

\begin{table}[h]
\small
\centering
\begin{tabular}{lccc}
\toprule
\textbf{Noun} & \textbf{Human} & \textbf{Ngrams} & \textbf{Noun $\delta$-LL} \\
\midrule
model (person) & 0.806 & 0.523 & 8.06 \\
animal & 0.722 & 0.699 & 2.97 \\
jet & 0.583 & 0.562 & 7.27 \\
aircraft & 0.583 & 0.551 & 3.92 \\
fish & 0.569 & 0.467 & -4.08 \\
vehicle & 0.542 & 0.468 & 4.66 \\
bus & 0.542 & 0.394 & 0.537 \\
tank & 0.542 & 0.564 & -0.639 \\
plane & 0.528 & 0.565 & 11.1 \\
car & 0.528 & 0.565 & 3.83 \\
motorcycle & 0.514 & 0.184 & 5.11 \\
truck & 0.514 & 0.437 & 13.6 \\
SUV & 0.480 & 0.500 & -2.27 \\
tractor & 0.401 & 0.500 & 11.2 \\
\bottomrule
\end{tabular}
\caption{Nouns in \textbf{intr-patient} sentences with normalized human ratings $\geq$ 0.5, along with their subject ratio from Google Syntactic Ngrams and the average $\delta$-LL from nouns in isolation (\ref{section:exp1}). The average $\delta$-LL for ``patient'' nouns ranges from -15.7 to 13.6. Note that {\it model} was presented to annotators with a disambiguating word sense ({\it person}).}
\label{tab:human-patient-nouns}
\end{table}

\subsection{Verbs with vehicle objects}
The other class of nouns present in Table \ref{tab:human-patient-nouns}, which also happen to be the remaining nouns, are vehicles. With regards to the relationship between animacy and agentivity, prior work such as \citet{zaenen2004animacy} has noted that ``intelligent machinery'' (such as computers and robots) and vehicles also often act as animates (below humans and above inanimates).  Interestingly, nearly half of the examples that \texttt{davinci-003} gets wrong are sentences containing verbs with vehicle objects (\textit{This car/vehicle/SUV/tractor/etc. drives nicely, This jet/plane/aircraft/etc. flies smoothly}). In fact, the examples that \texttt{davinci-003} gets the ``most wrong'' (higher $LL_{incorrect}-LL_{correct}$) are sentences with these verb-noun combinations.

Like the above examples with ``model'', some of these sentences have a possible alternative reading and are more ambiguous compared to sentences with verbs like {\it sell} (as in, {\it This car sells well.}). More specifically, they have a possible (though also less salient) unergative reading: e.g. in {\it This jet flies smoothly}, it could be a statement about how the jet flies on its own as opposed to about how the jet flies when someone flies it. Out of all the sentences in the test set, these are the only ones (along with some sentences with ``turn'') where the \textbf{intr-agent} has this possible unergative reading.

\section{Related Works}
There has been extensive work in the psycholinguistics literature investigating how humans make use of the relationship between events described by verbs and nouns that may participate in these events, which is especially relevant to the analysis described in \S\ref{section:animacy-thematic}. Works such as \citet{tanenhaus-1989} and \citet{trueswell-1994} have shown that humans utilize information about thematic fit to resolve ambiguity in sentence processing, mainly focusing on garden-path sentences. 

Along this line of work, \citet{mcrae-1998} and \citet{pado2007integration} created human judgement datasets for thematic fit by asking humans to rate nouns associated with events (e.g. a crook arresting/being arrested by someone) on a scale from 1 (very uncommon/implausible) to 7 (very common/plausible). As stimuli, humans are given the noun, the verb describing the event, and the role of the noun. While this setup is similar to our dataset, they focus on the explicit relationship between the event and the noun, while our data is meant to focus on the relationship between the prototypical role of a noun (out of context) and its role in a controlled syntactic environment. Furthermore, as we would like the agent/patient distinction to be a minimal pair resulting changing the noun in an identical surface form, the sets of nouns and verbs between their studies and ours only partially overlap.

This study also follows a well-established line of work on LMs as psycholinguistic subjects (\citealt{futrell-etal-2019-neural, ettinger-2020-bert, Linzen_2021}, inter alia). A large portion of this work focuses on probing LMs for sensitivity to the well-formedness of sentences containing various syntactic structures such as subject-verb agreement \cite{linzen-etal-2016-assessing}, relative clauses (\citealt{gulordava-etal-2018-colorless, ravfogel2021counterfactual}), and filler-gap dependencies \cite{wilcox2018rnn}, among others. A closely-related work by \citet{papadimitriou-etal-2022-classifying} investigates how BERT classifies grammatical role of entities in non-prototypical syntactic positions, similar to our setup in Experiment 3. 

There have also been works on evaluating and probing LMs for semantic/pragmatic knowledge. \citet{ettinger-2020-bert} created a suite of tests drawn from human language experiments to evaluate commonsense reasoning, event knowledge, and negation. The COGS challenge \cite{kim-linzen-2020-cogs}, which contains related tests to ours with regards to argument alternation, tests for whether LMs can learn to generalize about passivization and unnacusative-transitive alternations in English. \citet{misra2022comps} test LMs for their ability to attribute properties to concepts and further test property inheritance. With regards to lexical semantics, \citet{vulic-etal-2020-probing} investigate how type-level lexical information from words in context is stored in models across six typologically diverse languages.

However, our work is distinct from both previous syntax- and semantics-focused probing and evaluation in its focus on the interactions between the aspects of meaning in individual lexical items with larger syntactic structures or constructions. Nevertheless, methodologies from these research areas have informed the construction of our experiments. Our use of minimal pairs to form sentences with contrasting semantic roles is similar to the construction of the BLiMP dataset \cite{warstadt-etal-2020-blimp-benchmark} and other test suites. Furthermore, we treat the ``agent''/``patient'' labelling task as classification based on the generation probabilities of the labels, following \citet{linzen-etal-2016-assessing}'s method of using generation probabilities for grammaticality judgements. 

Another relevant recent line of work within NLP is inspired by Construction Grammar (CxG), a branch of theories within cognitive linguistics that posits that {\it constructions}---defined as form-meaning pairings---are the basic building blocks of language (\citealt{adele1995constructions, croft2001radical}, inter alia). \citet{mahowald2023discerning} conducted a similar prompting experiment on the English Article-Adjective-Numeral-Noun construction, though this was focused on grammaticality judgements as opposed to aspects of semantics. \citet{weissweiler2022} probe for both syntactic and semantic understanding of the English comparative correlative. Our study differs in that we analyze the impact of individual lexical items in what otherwise appears to be an identical syntactic construction, as opposed to analyzing competence of the construction as a whole. Finally, \citet{li-etal-2022-neural} find that sentences sharing the same argument structure constructions (ASCs) are closer in the embedding space than those sharing the main verb; in light of our results, an interesting direction would be to see if sentences of the same surface construction may cluster based on finer-grained semantic distinctions.

One consequence of our work---specifically with regards to \texttt{davinci-003}'s extremely high correlation with human judgements---is the potential for LMs as a tool for discovery in theoretical linguistics. This also has been argued recently by \citet{Petersen:Potts:2022}, who demonstrate this in the realm of lexical semantics through a case study of the English verb {\it break}. 

\section{Conclusion}
In order to gain insight into the behavior of LMs with respect to the syntax-semantics interface, we created a suite of prompting experiments focusing on agentivity. We prompt varying sizes of BLOOM, GPT-2, and GPT-3 to see if they are sensitive to aspects of agentivity at the lexical level, and then to see if they can either utilize or discard these word-level priors given the appropriate syntactic context. GPT-3 \texttt{davinci-003} performs exceptionally well in all three of our experiments---outperforming all other models tested by far---and is even better correlated with human judgements than some proxy corpus statistics. We find it surprising that \texttt{davinci-003} is able to capture an abstract notion of agentivity extremely well, but this ability does not appear to come from the size of the model alone as performance does not increase monotonically across any of the model families tested. What aspects of model training/data contribute to \texttt{davinci-003}'s (or other models') performance on linguistic tasks may be an interesting area for future work.

Furthermore, a qualitative analysis of what \texttt{davinci-003} gets incorrect reveals examples involving a number of linguistic confounders that make them more ambiguous to humans as well. The model's ability to ``pick out'' these linguistically interesting examples, combined with the high correlation with human ratings in Experiment 1, showcases the potential of LMs as tools for linguistic discovery for new phenomena, such as finding new classes of words or syntactic constructions that behave in unexpected ways. We hope these results encourage a more lively discussion between NLP researchers and linguists to unlock the potential of LMs as tools for theoretical linguistics research.

\section{Limitations}
While the use of a particular subset of English transitive verbs allows us to have precise control over the surface forms we are evaluating LMs on, this restricts our scope to a specific alternation in one language as well as a relatively small evaluation set. Nevertheless, we hope the methodology presented in this work can be extended to other phenomena across languages.

Additionally, while we explored a variety of ways to prompt these models, it may be the case that the prompt is non-optimal and therefore does not elicit the best possible output with respect to the task. Furthermore, the ``prompt'' to elicit human judgements is not the same as the prompt given to models, nor are the output formats (humans are asked to respond on a discrete scale from 1-5, while models are evaluated by their label log likelihoods). Evaluating whether the methodology in this line of work is a fair comparison between models and humans 
may be an interesting direction for future work.

\section{Acknowledgements}
This work was supported by NSF CISE RI grant number 2211951. For helpful comments and suggestions, we thank Kyle Mahowald and John Beavers, as well as the three anonymous reviewers and meta-reviewer. Finally, we would like to thank Thomas Lu, Saujas Vaduguru, Leilani Zhang, Russell Emerine, Vijay Viswanathan, Jeremiah Milbauer, Qianli Ma, Alex Wilf, Amanda Bertsch, Sireesh Gururaja, Yingshan Chang, Clara Na, Zhiruo Wang, Cathy Jiao, Simran Khanuja, Atharva Kulkarni, Anubha Kabra, Leena Mathur, and Shuyan Zhou for their help in annotating nouns in our dataset.

\bibliography{anthology,custom}

\appendix
\section{Noun-Verb-Adverb Combinations} \label{appendix:noun-verb-adv}

\begin{tabular}{c|p{0.7\linewidth}}
     verb & {\it \textbf{sells}} \\
     \multirow{2}{*}{nouns} & patients: {\it toy, book, novel, magazine, hat, lotion, album, car, SUV, product, make, item, CD, drug, snack} \\
     & agents: {\it salesman, saleswoman, businessman, businesswoman, trader, peddler, telemarketer, dealer, shopkeeper} \\
     adverbs & {\it easily, well, quickly} \\
     \midrule
     verb & {\it \textbf{drives}} \\
     \multirow{2}{*}{nouns} & patients: {\it car, SUV, truck, convertible, vehicle, tank, bus, tractor, van} \\
     & agents: {\it driver, person, chauffeur} \\
     adverbs & {\it nicely, smoothly, well} \\
     \midrule
     verb & {\it \textbf{flies}} \\
     \multirow{2}{*}{nouns} & patients: {\it plane, kite, jet, aircraft} \\
     & agents: {\it pilot, person, aviator, captain} \\
     adverbs & {\it nicely, smoothly, well} \\
     \midrule
     verb & {\it \textbf{cooks}} \\
     \multirow{2}{*}{nouns} & patients: {\it mushroom, pepper, fish, salmon, tuna, fillet, vegetable, herb, meat, ingredient, steak} \\
     & agents: {\it chef, cook, baker, caterer} \\
     adverbs & {\it nicely, well, terribly} \\
     \midrule
     verb & {\it \textbf{bakes}} \\
     \multirow{2}{*}{nouns} & patients: {\it pizza, potato, bread, cake, pastry, dough, pie, clay} \\
     & agents: {\it patissier, chef, cook, baker, person, confectioner} \\
     adverbs & {\it nicely, well, terribly} \\
     \midrule
     verb & {\it \textbf{reads}} \\
     \multirow{2}{*}{nouns} & patients: {\it passage, poem, verse, line, passage, script, abstract, essay, letter, report} \\
     & agents: {\it student, orator, person, narrator, announcer, broadcaster, teacher} \\
     adverbs & {\it nicely, well} \\
     \midrule
     verb & {\it \textbf{paints}} \\
     \multirow{2}{*}{nouns} & patients: {\it wall, fabric, glass, canvas, wood, surface, panel} \\
     & agents: {\it painter, artist, person, illustrator, portraitist} \\
     adverbs & {\it easily, terribly, well, beautifully} \\
\end{tabular}

\begin{tabular}{c|p{0.7\linewidth}}
     verb & {\it \textbf{writes}} \\
     \multirow{2}{*}{nouns} & patients: {\it section, passage, proposal, code, essay} \\
     & agents: {\it student, person, notetaker, journalist, scribe, doctor, professor, essayist, blogger, poet, novelist, author} \\
     adverbs & {\it quickly, easily} \\
    \midrule
    verb & {\it \textbf{performs}} \\
     \multirow{2}{*}{nouns} & patients: {\it routine, song, choreography, sonata, concerto, scene} \\
     & agents: {\it musician, person, actor, comedian, dancer, singer, soloist} \\
     adverbs & {\it easily} \\
    \midrule
    verb & {\it \textbf{photographs}} \\
     \multirow{2}{*}{nouns} & patients: {\it building, animal, landscape, lake, mountain, model, view} \\
     & agents: {\it photographer, cameraman} \\
     adverbs & {\it nicely, beautifully} \\
    \midrule
    verb & {\it \textbf{plays}} \\
     \multirow{2}{*}{nouns} & patients: {\it cello, piano, violin, instrument, flute, clarinet} \\
     & agents: {\it musician, violinist, cellist, pianist, drummer, flutist, clarinetist} \\
     adverbs & {\it nicely, beautifully} \\
    \midrule
    verb & {\it \textbf{cuts}} \\
     \multirow{2}{*}{nouns} & patients: {\it meat, cardboard, packaging, board, paper, fabric} \\
     & agents: {\it hairdresser, barber, butcher, chef} \\
     adverbs & {\it nicely, roughly, cleanly, effortlessly} \\
    \midrule
    verb & {\it \textbf{cleans}} \\
     \multirow{2}{*}{nouns} & patients: {\it jewelry, window, countertop, floor, surface, carpet, windshield, mirror, pot, silverware, bedding} \\
     & agents: {\it janitor, maid, cleaner, housekeeper, busboy, waiter, waitress} \\
     adverbs & {\it easily, quickly, effortlessly} \\
    \midrule
    verb & {\it \textbf{washes}} \\
     \multirow{2}{*}{nouns} & patients: {\it bottle, tub, shirt, car, windshield, dish, bedding, blanket, bowl} \\
     & agents: {\it worker, maid, cleaner, busboy} \\
     adverbs & {\it easily, quickly} \\
    \midrule
    verb & {\it \textbf{shaves}} \\
     \multirow{2}{*}{nouns} & patients: {\it beard, stubble, sideburn} \\
     & agents: {\it barber, hairdresser} \\
     adverbs & {\it neatly, nicely, smoothly} \\
\end{tabular}

\noindent \begin{tabular}{c|p{0.75\linewidth}}
     verb & {\it \textbf{packs}} \\
     \multirow{2}{*}{nouns} & patients: {\it crate, lunchbox, basket, container, coat, jacket, bag, duffle, food, suitcase, tent, backpack} \\
     & agents: {\it mover, traveller, clerk, worker, backpacker, roadtripper, hiker, camper} \\
     adverbs & {\it well, easily} \\
     \midrule
     verb & {\it \textbf{stitches}} \\
     \multirow{2}{*}{nouns} & patients: {\it silk, quilt, cotton, cut, cloth, fabric, wound} \\
     & agents: {\it surgeon, tailor, machine, upholsterer, dressmaker} \\
     adverbs & {\it easily, smoothly, nicely, poorly} \\
     \midrule
     verb & {\it \textbf{embroiders}} \\
     \multirow{2}{*}{nouns} & patients: {\it cushion, thread, cloth, fabric} \\
     & agents: {\it tailor, seamster, seamstress} \\
     adverbs & {\it well, nicely, beautifully, poorly} \\
     \midrule
     verb & {\it \textbf{knits}} \\
     \multirow{2}{*}{nouns} & patients: {\it yarn, wool, pattern} \\
     & agents: {\it person, lady, man, woman} \\
     adverbs & {\it well, nicely, beautifully, poorly, easily} \\
     \midrule
     verb & {\it \textbf{sews}} \\
     \multirow{2}{*}{nouns} & patients: {\it fabric, material} \\
     & agents: {\it tailor, seamster, machine} \\
     adverbs & {\it well, nicely, beautifully, poorly} \\
     \midrule
     verb & {\it \textbf{turns}} \\
     \multirow{2}{*}{nouns} & patients: {\it screw, knob, car, bike, motorcycle, valve, handle} \\
     & agents: {\it driver, racer, motorist, pilot} \\
     adverbs & {\it smoothly, easily, nicely, roughly} \\
     \midrule
     verb & {\it \textbf{carves}} \\
     \multirow{2}{*}{nouns} & patients: {\it pumpkin, wood, stone, gem, ice, steak, turkey} \\
     & agents: {\it sculptor, person, jeweler, artisan, carver} \\
     adverbs & {\it beautifully, nicely, cleanly, flawlessly} \\
     \midrule
     verb & {\it \textbf{sculpts}} \\
     \multirow{2}{*}{nouns} & patients: {\it wood, stone, marble, ice, clay} \\
     & agents: {\it sculptor, person, potter, mason, carver} \\
     adverbs & {\it beautifully, nicely, cleanly} \\
\end{tabular}

\section{Data Curation Criteria} \label{appendix:data-curation}
After collecting a list of optionally transitive verbs that appear as intransitive via object drop (agent subject) or object promotion in the form of the middle construction (patient subject), we then had to curate adverbs and nouns that work in the templates as described in Table \ref{tab:templates}.

Adverbs must be manner adverbs, but they should not be {\it agent-oriented} adverbs (\citealt{jackendoff1972semantic, ernst2001syntax}) that express the mental state of the agent. Examples of such adverbs include {\it furiously, happily, angrily}, etc. 

Then for each verb and a list of adverbs for each verb, we come up with a list of patient and agent nouns. All of the nouns must work in intransitive and transitive templates using the same sense of the verb. For nouns added as patients in the intransitive, the noun must not be an entity that causes the event described by the verb. Furthermore, it should not be necessarily oblique in the transitive form. In the example below, {\it needle} cannot be the direct object of the transitive and can only appear in the {\it with} prepositional phrase, so we do not include it in the list of nouns:

\begin{enumerate}[resume]
\item 
        \begin{enumerate}[label=\alph*., ref=\alph*]
        \itemsep0em 
            \item This needle sews easily.
            \item The tailor sews easily with this needle.
            \item *The tailor sews this needle easily.
        \end{enumerate}
\end{enumerate}

For nouns added as agents, in the intransitive it must be clear that the noun is the one doing the action. For human agents, we try to add agents that are most closely associated to the action described for the event, especially with those that tend to take human direct objects in the transitive form, such as {\it shave}.

\section{Human Annotation Details} \label{appendix:human}
We had 19 human annotators rate all 233 unique nouns on Google Forms. Each annotator saw a different random order of the nouns and were presented with 10 nouns on each page of the form, though they could go back to alter previous responses. All annotators are fluent in English. Annotators were also asked to self-identify as native or non-native speakers; 14 of 19 consider themselves native speakers.

For nouns that have multiple common and highly distinct word senses, we gave annotators a short disambiguating description. This description does not contain any verbs or any other indicator for what types of events the entity may occur in. A list of these nouns with their disambiguating description is given in Table \ref{tab:senses}.

\begin{table}[h]
\small
\centering
\begin{tabular}{cp{0.75\linewidth}}
\toprule
\textbf{Noun} & \textbf{Description} \\
\midrule
{\it make} & product of a particular company, such as of a car \\
{\it plane} & airplane \\
{\it kite} & a light frame covered with paper, cloth, or plastic, often with a stabilizing tail \\
{\it jet} & aircraft \\
{\it line} & of a text/a poem/etc. \\
{\it passage} & of a text/an essay/etc. \\
{\it panel} & of wood/a hard surface/etc. \\
{\it model} & person \\
{\it routine} & a part of an entertainment act \\
{\it board} & a long, thin, flat piece of wood or other hard material \\
{\it letter} & a sheet of paper with words on it in an envelope \\
{\it proposal} & a formal plan or suggestion \\
{\it turkey} & meat \\
\bottomrule
\end{tabular}%
\caption{Nouns and disambiguating descriptions given to annotators.}
\label{tab:senses}
\end{table}

\subsection{Instructions provided to annotators}
An \textbf{agent} is something that initiates an action, possibly with some degree of volition. In other words, nouns that tend to be agents have a tendency to do things. \\

\noindent A \textbf{patient} is something that undergoes an action and often experiences a change. In other words, nouns that tend to be patients have a tendency to have things done to it. \\

\noindent In this form, you are tasked to annotate how "agentive" you think a noun typically is---in other words, how likely it is to be an agent or a patient when an action involving both an agent and a patient occur. \\

\noindent Ex: The plant was watered by John. \\
\noindent The plant = patient \\
\noindent John = agent \\

\noindent Ex: The sun burns John. \\
\noindent The sun = agent \\
\noindent John = patient \\ 

\noindent A more formal definition is given by Dowty (1991), who outlines contributing properties of agents and patients:
\begin{enumerate}
    \item Contributing properties for the Agent Proto-Role: 
    \begin{itemize}
    \itemsep0em 
        \item volitional involvement in the event or state — sentience (and/or perception)
        \item causing an event or change of state in another participant
        \item movement (relative to the position of another participant)
        \item (exists independently of the event named by the verb)
    \end{itemize}
    \item Contributing properties for the Patient Proto-Role:
    \begin{itemize}
    \itemsep0em 
        \item undergoes change of state
        \item incremental theme (something that changes incrementally over the course of an event)
        \item causally affected by another participant
        \item stationary relative to movement of another participant
        \item (does not exist independently of the event, or not at all)
    \end{itemize}
\end{enumerate}

\noindent For the sake of simplicity, disregard events described by reflexives (such as John shaved himself).

\noindent For each of the following nouns, rate it on the following scale: \\

\noindent 1 = very unlikely to be an agent \\
\noindent 2 = somewhat unlikely to be an agent \\
\noindent 3 = no preference between agent and patient \\
\noindent 4 = somewhat likely to be an agent \\
\noindent 5 = very likely to be an agent \\

\begin{figure}[h]
\centering
\includegraphics[width=\linewidth]{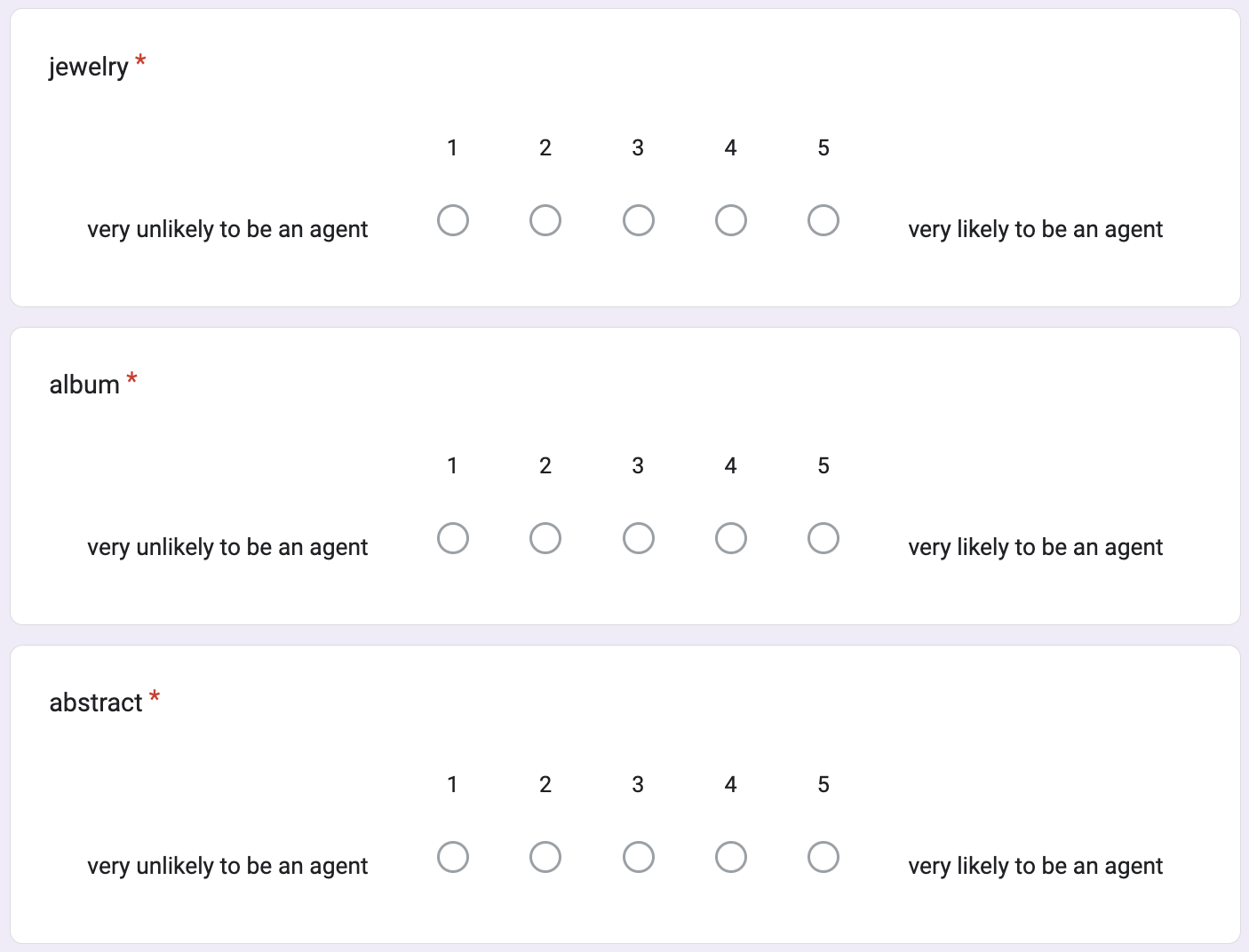}
\caption {Example of Google Form question format given to annotators.
}
\end{figure}

\section{Results by Example Order} \label{appendix:order}
Tables \ref{tab:ordering-exp1}, \ref{tab:ordering-exp2}, and \ref{tab:ordering-exp3} show performance in both APAP and PAPA orderings in Experiments 1 (nouns in isolation), 2 (nouns in intransitive sentences), and 3 (nouns in transitive sentences) respectively. For simplicity, we only report correlations with human judgements.

\begin{table}[t]
\centering
\begin{tabular}{@{}lrrr@{}}
\toprule
\textbf{Model}   & \multicolumn{1}{c}{\textbf{APAP}} & \multicolumn{1}{c}{\textbf{PAPA}} & \multicolumn{1}{c}{$\delta$} \\ \midrule
BLOOM \texttt{560m}       & 0.566                             & 0.531                             & 0.036                              \\
BLOOM \texttt{1b1}        & 0.384                             & 0.365                             & 0.019                              \\
BLOOM \texttt{1b7}        & 0.308                             & 0.371                             & 0.062                              \\
BLOOM \texttt{3b}         & 0.476                             & 0.133                             & 0.343                              \\
BLOOM \texttt{7b1}        & -0.118                            & 0.150                             & 0.268                              \\ 
\midrule
GPT-2 \texttt{small}       & 0.648                             & 0.652                             & 0.004                              \\
GPT-2 \texttt{medium}      & 0.420                             & 0.367                             & 0.053                              \\
GPT-2 \texttt{large}       & 0.501                             & 0.496                             & 0.005                              \\
GPT-2 \texttt{xl}          & 0.486                             & 0.231                             & 0.255                              \\
\midrule
GPT-3 \texttt{ada-001}     & 0.589                             & 0.598                             & 0.009                              \\
GPT-3 \texttt{babbage-001} & 0.394                             & 0.228                             & 0.166                              \\
GPT-3 \texttt{curie-001}   & 0.418                             & -0.204                            & 0.622                              \\
GPT-3 \texttt{davinci-001} & 0.579                             & 0.356                             & 0.223                              \\
GPT-3 \texttt{davinci-003} & \textbf{0.934}                    & \textbf{0.943}                    & 0.010                              \\ \bottomrule
\end{tabular}
\caption{\textbf{Experiment 1}: Correlation between the difference in log-likelihood of predicting “agent” or “patient” with human ratings for nouns in isolation in both example orderings.}
\label{tab:ordering-exp1}
\end{table}

\begin{table}[t]
\centering
\begin{tabular}{@{}lrrr@{}}
\toprule
\textbf{Model}   & \multicolumn{1}{c}{\textbf{APAP}} & \multicolumn{1}{c}{\textbf{PAPA}} & \multicolumn{1}{c}{$\delta$} \\ \midrule
BLOOM \texttt{560m}       & 0.214                             & 0.219                             & 0.005                              \\
BLOOM \texttt{1b1}        & -0.096                            & 0.027                             & 0.124                              \\
BLOOM \texttt{1b7}        & 0.618                             & 0.795                             & 0.177                              \\
BLOOM \texttt{3b}         & 0.049                             & 0.512                             & 0.463                              \\
BLOOM \texttt{7b1}        & 0.050                             & 0.272                             & 0.223                              \\
\midrule
GPT-2 \texttt{small}       & 0.658                             & 0.190                             & 0.468                              \\
GPT-2 \texttt{medium}      & 0.546                             & 0.500                             & 0.047                              \\
GPT-2 \texttt{large}       & 0.632                             & 0.586                             & 0.045                              \\
GPT-2 \texttt{xl}          & 0.484                             & 0.531                             & 0.047                            
 \\
\midrule
GPT-3 \texttt{ada-001}     & 0.574                             & 0.417                             & 0.157                              \\
GPT-3 \texttt{babbage-001} & -0.030                            & -0.322                            & 0.292                              \\
GPT-3 \texttt{curie-001}   & 0.045                             & 0.266                             & 0.221                              \\
GPT-3 \texttt{davinci-001} & 0.673                             & 0.622                             & 0.051                              \\
GPT-3 \texttt{davinci-003} & \textbf{0.927}                    & \textbf{0.911}                    & 0.017                              \\ \bottomrule
\end{tabular}
\caption{\textbf{Experiment 2}: Correlation between the difference in log-likelihood of predicting “agent” or “patient” with human ratings for nouns in intransitive sentences in both example orderings.}
\label{tab:ordering-exp2}
\end{table}

\begin{table*}[t]
\centering
\begin{tabular}{@{}lrrrrrr@{}}
\toprule
\textbf{}         & \multicolumn{3}{c}{\textbf{trans-agent}}                                                                   & \multicolumn{3}{c}{\textbf{trans-patient}} \\
\cmidrule(lr){2-4}\cmidrule(lr){5-7}
\textbf{Model}    & \multicolumn{1}{c}{\textbf{APAP}} & \multicolumn{1}{c}{\textbf{PAPA}} & \multicolumn{1}{c}{$\delta$} & \multicolumn{1}{c}{\textbf{APAP}} & \multicolumn{1}{c}{\textbf{PAPA}} & \multicolumn{1}{c}{$\delta$} \\
\midrule
BLOOM \texttt{560m}        & 0.034                             & 0.090                             & 0.056                              & 0.962                             & 0.932                             & 0.031                              \\
BLOOM \texttt{1b1}         & 0.620                             & 0.781                             & 0.161                              & 0.516                             & 0.457                             & 0.059                              \\
BLOOM \texttt{1b7}         & 0.940                             & 0.013                             & 0.927                              & 0.007                             & 0.989                             & 0.982                              \\
BLOOM \texttt{3b}          & 1.000                             & 0.059                             & 0.941                              & 0.000                             & 0.895                             & 0.895                              \\
BLOOM \texttt{7b1}         & 0.974                             & 0.017                             & 0.957                              & 0.088                             & 1.000                             & 0.912                              \\
\midrule
GPT-2 \texttt{small}      & 0.313                             & 0.796                             & 0.483                              & 0.811                             & 0.210                             & 0.600                              \\
GPT-2 \texttt{medium}      & 0.121                             & 0.000                             & 0.121                              & 0.877                             & 1.000                             & 0.123                              \\
GPT-2 \texttt{large}       & 0.829                             & 0.389                             & 0.440                              & 0.163                             & 0.623                             & 0.461                              \\
GPT-2 \texttt{xl}          & 0.978                             & 0.001                             & 0.977                              & 0.018                             & 1.000                             & 0.982                              \\
\midrule
GPT-3 \texttt{ada-001}     & 0.313                             & 0.089                             & 0.224                              & 0.611                             & 0.933                             & 0.322                              \\
GPT-3 \texttt{babbage-001} & 0.987                             & 0.044                             & 0.943                              & 0.023                             & 0.994                             & 0.971                              \\
GPT-3 \texttt{curie-001}   & 0.353                             & 0.034                             & 0.319                              & 0.740                             & 0.963                             & 0.224                              \\
GPT-3 \texttt{davinci-001} & 0.987                             & 0.968                             & 0.018                              & 0.413                             & 0.427                             & 0.013                              \\
GPT-3 \texttt{davinci-003} & 0.996                             & 0.993                             & 0.004                              & 0.999                             & 0.984                             & 0.015                              \\   
\bottomrule
\end{tabular}
\caption{\textbf{Experiment 3}: Accuracy in both example orderings for predicting the role of the noun in transitive sentences, where \textbf{trans-agent} corresponds to the noun in subject position and \textbf{trans-patient} to object position.}
\label{tab:ordering-exp3}
\end{table*}

Both GPT-3 \texttt{davinci-001} and \texttt{davinci-003} are very robust to changes in example ordering for all three experiments, as are BLOOM \texttt{560m} and \texttt{1b1}. The three largest BLOOM models are remarkably sensitive to ordering, especially in Experiment 3, as are GPT-2 \texttt{xl} and GPT-3 \texttt{curie-001} and \texttt{babbage-001}.  

\section{Propbank Statistics} \label{appendix:propbank}
When calculating model correlations with Propbank, we use all nouns with at least one occurrence of appearing within an ARG0/1 span in the parse tree to maximize the number of nouns we can compare with. However, we recognize that this may mess with correlation values since nouns with only one occurrence will have values at either 0 or 1. Furthermore, depending on the role the noun has in that particular sentence, it may push its agent rating to the opposite end of the spectrum compared to its ``typical'' behavior. Thus, we also tried calculating the correlation only for nouns that occur some greater number of times (within an ARG0/1 span) in Propbank. We call the minimum number of times the noun must appear the \textbf{count threshold}.

\begin{figure}[h]
\centering
\includegraphics[width=\linewidth]{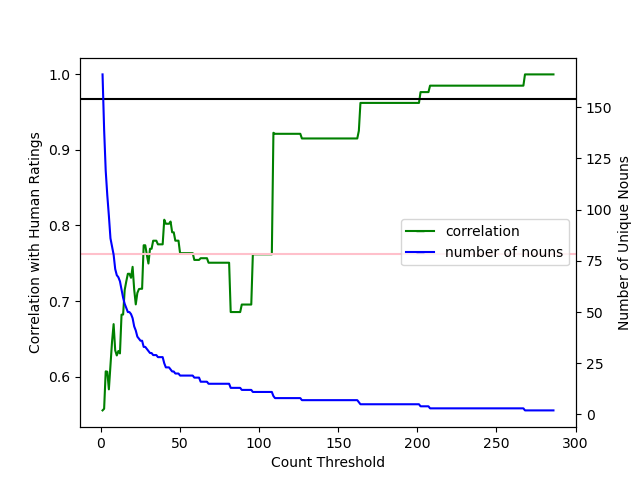}
\caption{\label{fig:propbank} Count threshold versus the correlation between noun agent ratios and human ratings and the number of unique nouns that surpass the threshold. The pink horizontal line shows the correlation of Google Syntactic Ngrams with human ratings; the black line shows the average inter-annotator group correlation.
}
\end{figure}

Figure \ref{fig:propbank} plots the Propbank agent ratio correlation with human ratings against the count threshold (in green). We also plot the number of nouns that meet this count threshold (in blue). The minimum count threshold to have a greater correlation than Google Syntactic Ngrams (pink line) is 27, however only 33 nouns meet this threshold. To meet meet the average human inter-annotator group correlation, the threshold is 268; only two nouns meet this. 

\section{Adjusting Threshold for Exp 2}
\begin{figure}[h]
\centering
\includegraphics[width=\linewidth]{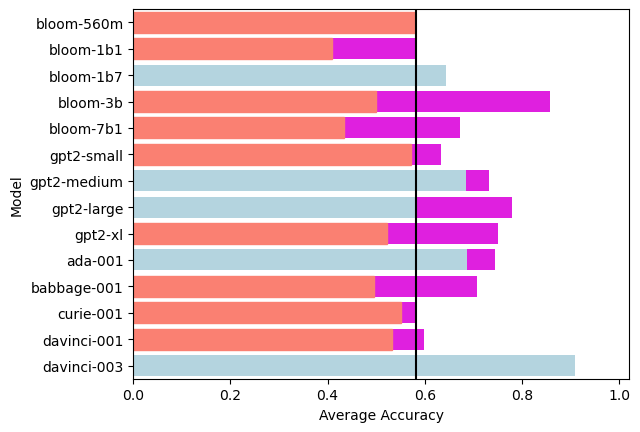}
\caption{\label{fig:intr_thresh_acc} Average accuracy for predicting the label in \textbf{intr-agent}/\textbf{intr-patient} sentences with adjusted thresholds. After this adjustment, all models are at or above majority class accuracy. Magenta segments show increase in performance.
}
\end{figure}

We also considered the possibility that the models may have a bias towards either the ``agent'' or ``patient'' label and may actually be correctly classifying nouns given an appropriate non-zero threshold for $\delta$-LL. To account for this, we recalculate accuracies with thresholds that provide the best performance for each model as an ``upper bound'' for performance, as seen in Figure \ref{fig:intr_thresh_acc}. After this adjustment, all models do at least as well as predicting the majority class, with GPT-2 \texttt{xl} experiencing the largest gain in accuracy. Nevertheless, GPT-3 \texttt{davinci-003} still outperforms all other models by far.

\end{document}